# HAIDA: Biometric technological therapy tools for neurorehabilitation of Cognitive Impairment

E. Fernandez, J. Solé-Casals, P. M. Calvo, M. Faundez-Zanuy, K. Lopez-de-Ipina

*Abstract*— Dementia, and specially Alzheimer's disease (AD) and Mild Cognitive Impairment (MCI) are one of the most important diseases suffered by elderly population. Music therapy is one of the most widely used non-pharmacological treatment in the field of cognitive impairments, given that music influences their mood, behavior, the decrease of anxiety, as well as facilitating reminiscence, emotional expressions and movement. In this work we present HAIDA, a multi-platform support system for Musical Therapy oriented to cognitive impairment, which includes not only therapy tools but also non-invasive biometric analysis, speech, activity and hand activity. At this moment the system is on use and recording the first sets of data. Results obtained using HAIDA will be presented in a near future after the analysis.

## I. INTRODUCTION

Alzheimer's Disease (AD) is characterized by a progressive and irreversible cognitive deterioration along with other deficits and behavioral symptoms. Its prevalence keeps increasing mainly in elderly people, and as highlighted by the World Alzheimer Reports, AD is becoming epidemic as 900 million people can be considered as the world's elderly population, while most of them live in developed countries [1]. Therefore, an early and accurate diagnosis of AD could help patients and their families to plan the future, and offers the best possibilities of symptoms being treated. Previously in an early stage, the cognitive loss or Mild Cognitive Impairment (MCI) appears, but it does not seem sufficiently severe to interfere in independent abilities of daily life, so it usually does not receive an appropriate diagnosis, and later some of them develop AD. The detection of MCI is a challenging issue to be addressed by medical specialists and could help future AD patients [2]. Along with memory loss, other problems appear such as the loss of social and language skills, emotion, expression and/or changes in handwriting and drawing. These inabilities to communicate and to develop everyday activities appear in early stages leads to their social exclusion, and to a serious negative impact not only on the sufferers, but also on their relatives [3]. In the last years several works in the state of the art have addressed the automatic analysis of these elements [2, 4]. In the present work we focus on a novel approach by the integration of more robust and multimodal tools to extract, integrate and correlate non-invasive biomarkers from oral expression (speech, disfluencies) and handwriting/drawing (on-surface and in-air) signals [5, 6]. Machine Learning and Deep Learning Paradigms will be used for modeling, as well as several feature-sets based on linear and non-linear approaches in order to develop a real-time and robust system.

## II. MUSIC THERAPY

According to the World Federation of Music Therapy (WFMT), "Music therapy consists in the use of music and/or its elements (sound, rhythm, melody, harmony) by a music therapist, with a patient or group, in the process and promote communication, learning, mobilization, expression, organization, or other relevant therapeutic objectives, in order to achieve changes and satisfy physical, emotional, mental, social and cognitive needs. Music therapy seeks to discover potentials and/or restore functions of the individual so that it reaches a better intra and/or interpersonal organization and, consequently, a better quality of life through prevention and rehabilitation in a treatment". Music therapy, together with cognitive stimulation, are the two most widely used non-pharmacological treatments in the field of cognitive impairments. The practice of music therapy with people with cognitive impairment has allowed to study that music influences their mood, behavior, the decrease of anxiety, as well as facilitating reminiscence, emotional expressions and movement. As in cognitive stimulation, it is very important that the material used to perform the intervention is significant for elicitation, in the case of music, if we take into account their tastes and experiences, we will obtain better results. The term "music therapy", defined by the World Federation of Music Therapy [7], is a health discipline registered as such. There are reviews of the music therapy literature in the field of dementia in which its effects on symptomatic reduction are discussed [8-10]. However, although the term "music therapy" is used in these reviews, some of the studies included are actually studies of "musical interventions" that employ group or individual activities and incorporate listening activities and exercises with recorded music. and the follow-up of melodies with instruments or with the voice

This work has been done in collaboration with AFAGI, partially supported by the Basque Goverment, the University of the Basque Country by the IT11156 project – ELEKIN, Diputación Foral de Gipuzkoa, University of Vic – Central University of Catalonia under the research grant R0947, and the Spanish Ministry of Science and Innovation TEC2016-77791-C04-RX.

P.M. Calvo, E. Fernandez and K. Lopez-de-Ipina are with the University of the Basque Country UPV/EHU, Donostia, Spain; (corresponding author to provide e-mail: karmele.ipina @ehu.eus).

J. Solé-Casals is with the Data and Signal Processing Research Group, University of Vic – Central University of Catalonia, Vic, Catalonia, Spain.

M. Faundez-Zanuy is with the Escola Superior Politècnica Tecnocampus - Universitat Pompeu Fabra, Mataró, Catalonia, Spain.

[11-15]. Recent results from Ridder [16] indicate that interactive live music therapy reduces agitation, anxiety and the need for medication. Research on the interactive elements needed in an individual music therapy is not very extensive [10]. There is currently no consensus among researchers about how music therapy is related in a broader context to patient care [17, 18].

## III. HAIDA

HAIDA is a multi-platform support system for Musical Therapy oriented to cognitive impairment. HAIDA includes not only therapy tools but also non-invasive biometric analysis, speech, activity and hand activity (see Fig. 1). The main objectives of the system are the following:

- Create a customizable therapy tool for each person depending on the stage of dementia.
- Analyze, through Deep-Learning techniques and Convolutive Neural Networks, the emotions of the person and different biometric data.
- To create a non-verbal communication tool that can be used in an intuitive, inexpensive and mobile interface, such as a tablet.
- Use non-verbal communication to adapt and personalize the tool of music therapy to each person.
- Have an objective biometric evaluation by means of automatic detection of people's emotions, both using the image of the face, the voice and body movements.

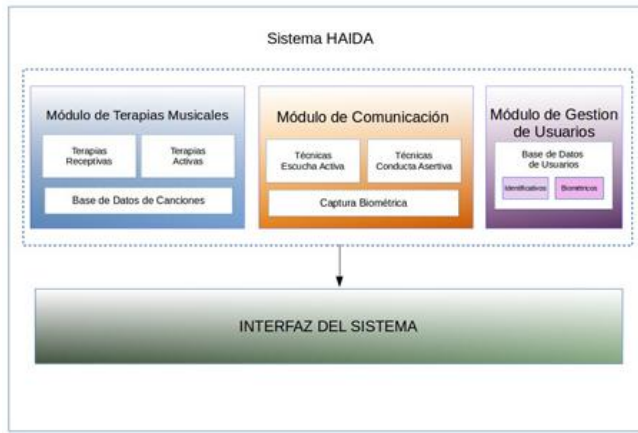

Fig. 1. Block diagram of the HAIDA system

## IV. DISCUSSION AND CONCLUSION

The role of non-verbal behavior in people with impairments offers enormous potential for caregivers and for the patients themselves, increasing the effect of non-pharmacological therapies, preserving their identity and improving their quality of life. Memory loss is a classic symptom in AD It has been shown that the more emotional a memory is, the more it looks at his/her memory. This is the case of memories associated with songs or the songs themselves. That is why professionals take advantage of this sensory capacity for a cognitive stimulation. The attention and emotion that emotionally charged family stimuli produce have been studied for each of the Alzheimer's disease patients. These investigations can lead to the implementation of individualized therapies promoting through concrete stimuli to each patient, a more specific sensory-cognitive stimulation. This project aims to be a breakthrough in music therapy tool and a new patient-caregiver communication system that can serve to improve the application of the various therapies, reduce stress and anxiety caused to a large extent by the patient's frustration at losing communication skills